\documentclass[10pt,twocolumn,letterpaper]{article}

\usepackage{iccv}
\usepackage{times}
\usepackage{epsfig}
\usepackage{graphicx}
\usepackage{amsmath}
\usepackage{amssymb}
\usepackage{bm}
\usepackage{booktabs}
\usepackage{caption}
\usepackage{subcaption}

\usepackage[pagebackref=true,breaklinks=true,letterpaper=true,colorlinks,bookmarks=false]{hyperref}

\iccvfinalcopy 

\def\iccvPaperID{898} 

\ificcvfinal\fi
\begin{document}

\title{Biphasic Learning of GANs for High-Resolution Image-to-Image Translation}

\author{Jie Cao$^{1}$, Huaibo Huang$^{1}$, Yi Li$^{1}$, Jingtuo Liu$^{2}$, Ran He$^{1}$, Zhenan Sun$^{1}$\\
$^1$CRIPAC \& NLPR, Institute of Automation, Chinese Academy of Sciences.\\
$^2$ Baidu, Inc. \\
{\tt\small \{jie.cao,huaibo.huang,yi.li\}@cripac.ia.ac.cn, \{rhe,znsun\}@nlpr.ia.ac.cn} \\
{\tt \small \{liujingtuo\}@baidu.com}
}

\twocolumn[{
\renewcommand\twocolumn[1][]{#1}
\maketitle
\begin{center}
    \includegraphics[width=\textwidth]{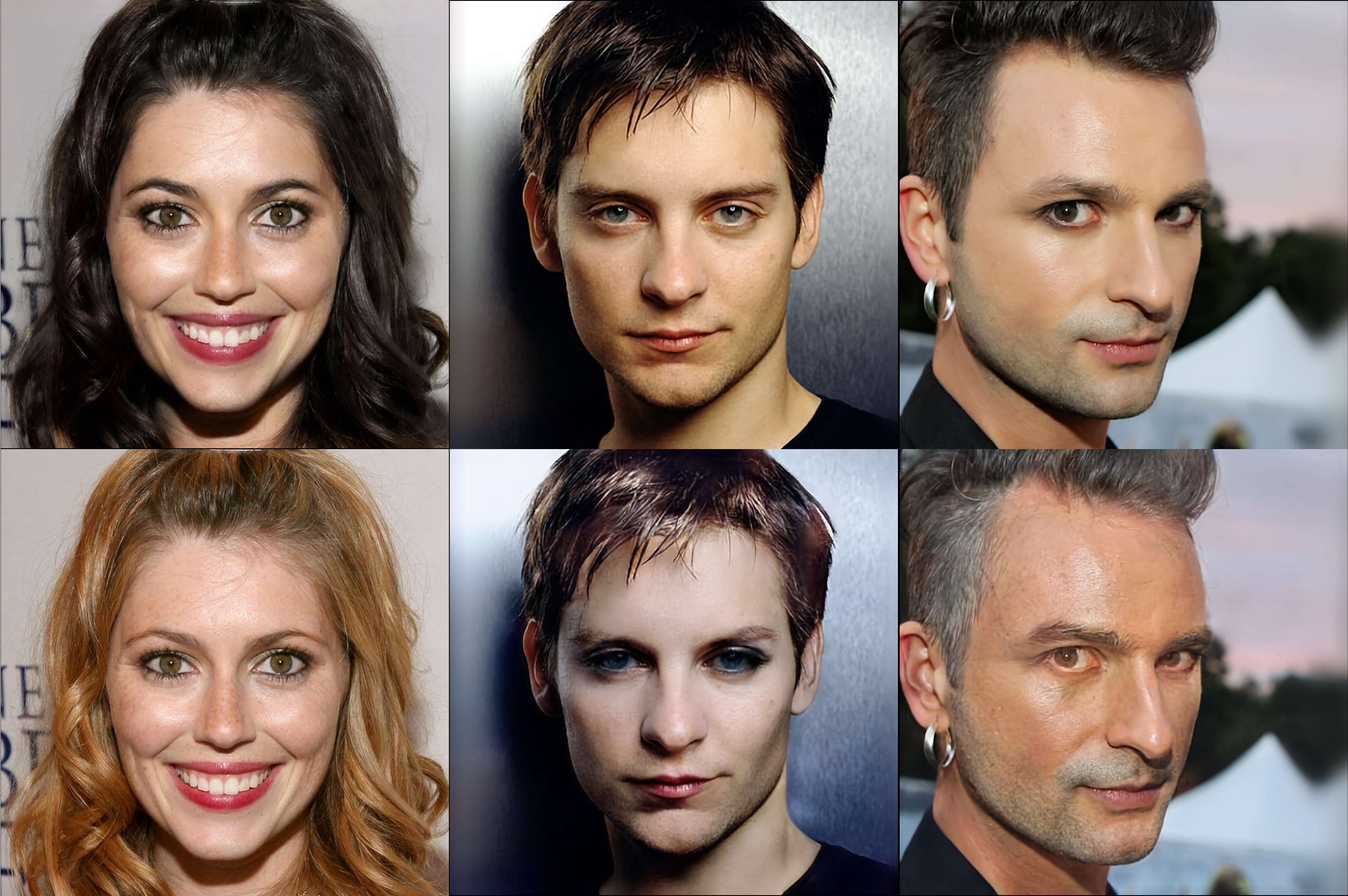}
    \captionof{figure}{Image-to-image translation results at $1024^2$ resolution. Given source images (the first row), our algorithm automatically produces transformed images (the second row) with specified attributes (blond hair, adding makeup, and aging). Due to the limitation of submission size, we have to downsample this figure.}
    \label{f1}
\end{center}
}]

\begin{abstract}
Despite that the performance of image-to-image translation has been significantly improved by recent progress in generative models, current methods still suffer from severe degradation in training stability and sample quality when applied to the high-resolution situation. In this work, we present a novel training framework for GANs, namely biphasic learning, to achieve image-to-image translation in multiple visual domains at $1024^2$ resolution. Our core idea is to design an adjustable objective function that varies across training phases. Within the biphasic learning framework, we propose a novel inherited adversarial loss to achieve the enhancement of model capacity and stabilize the training phase transition. Furthermore, we introduce a perceptual-level consistency loss through mutual information estimation and maximization. To verify the superiority of the proposed method, we apply it to a wide range of face-related synthesis tasks and conduct experiments on multiple large-scale datasets. Through comprehensive quantitative analyses, we demonstrate that our method significantly outperforms existing methods.
\end{abstract}

\section{Introduction}

Image-to-image translation is a classical vision problem whose goal is to learn the mapping function among multiple visual domains. Different from image generation from noise, the translation task requires a balanced approach that meets both the visual quality and overall consistency between the source and the transformed images. Recent progress in generative methods, including generative adversarial networks (GANs) \cite{goodfellow2014generative}, variational autoencoders (VAE) \cite{kingma2013auto}, and autoregressive models \cite{oord2016pixel}, have dramatically improved the quality of the transformed images. However, high-resolution image-to-image translation is far from being solved. When dealing with images at $1024^2$, the sample quality and training stability of current methods decline sharply, portending there is a long road to achieve true photorealism generation.

To push the frontiers of image-to-image translation to a higher resolution, we design a novel training method for GANs, namely biphasic learning, that improves the adversarial supervision manner. Our method has two distinct characteristics: 1) our generator and discriminator start joint adversarial training from dealing with images at a relatively low resolution in the initial phase, and then they are enhanced to produce full-resolution images in the enhancing phase; 2) the evaluation and optimization criterions of our loss items update with the training phase transition. We regard the transition from the initial to the enhancing training phase as a knowledge transfer process \cite{hinton2015distilling}. The discriminator in the former stage plays the role of teacher net and guides our network in the latter stage to inherit the learned knowledge. In the enhancing phase, we adjust the distillation objective to the adversarial learning condition. Consequently, we propose the inherited adversarial loss to guide the competing training process of the generator and the discriminator. Furthermore, in the initial phase, we borrow several auxiliary discriminators employed in normal-resolution translation tasks. In the enhancing phase, we fix these discriminators, turning the adversarial training into supervised training. Thanks to the effective adjustable supervision manner, our network is stable during the training phase transition and produces high-resolution samples of good quality.

We also propose a novel mutual perceptual information regularization to maintain the consistency between the source and the transformed images. Note that we mainly consider the translation tasks in the unconstrained situations where the paired training data is not available. Despite that existing methods, \eg, CycleGAN \cite{zhu2017unpaired}, have made great efforts, their sample quality in the high-resolution situation is still very limited. They mainly measure image similarity in the pixel domain, which leads to the neglect of fine-grained texture information. In the high-resolution situation, the over-smoothed results caused by the heavy reliance on the pixel-level loss have significant flaws in visual realism. To address this issue, we explore an alternative strategy and maximize the mutual information between the inputs and outputs. We employ a deep mutual information estimator trained on external data and design a supervised learning scheme to guide the estimator to measure image similarity in the perceptual level. Benefiting from the effective training method, our estimator is robust and more powerful in capturing the local texture information than existing unsupervised approaches \cite{brakel2017learning,belghazi2018mine,hjelm2018learning}. Equipped with the mutual perceptual information regularization, our network casts off the dilemma of either sample quality or visual consistency.

We verify the effectiveness of our approach on CelebA-HQ \cite{ karras2017progressive}. Furthermore, we also conduct comparison experiments on normal-resolution face datasets, including RaFD \cite{langner2010presentation} and Multi-PIE \cite{gross2010multi}. Experimental results demonstrate that our approach not only outperforms current state-of-the-art methods in quantitative measurements but also advances the visual appearance and the resolution of image-to-image translation.

\begin{figure*}[!tbp]
\begin{center}
\centering
\includegraphics[width=\textwidth]
{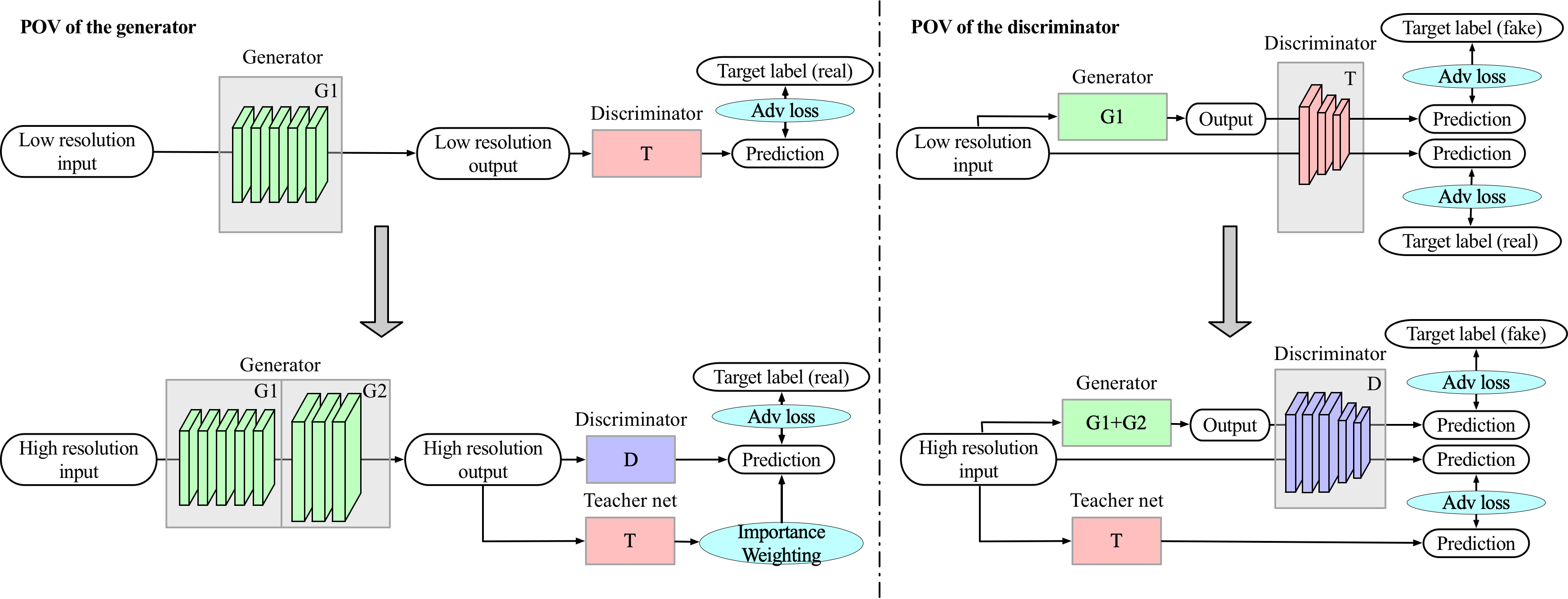}
\end{center}
\caption{An illustration of the proposed biphasic learning. The left and the right parts depict the network forward propagation in POV (point-of-view) of the generator and the discriminator, respectively. In the initial phase (shown in the upper part), the networks are trained in the classical adversarial training scheme. In the enhancing phase (shown in the lower part), importance weighting and knowledge distillation schemes are integrated into the training process with the aid of the discriminator trained in the former phase. The target domain label input is omitted for clarity.}
\label{f2}
\end{figure*}

\section{Related work}

\textbf{High-resolution image generation} is a broad and challenging study topic in computer vision and graphics. Particular study interest in this area has been raised along with the release of high-resolution datasets \cite{karras2017progressive, karras2018style}. In the field of image generation originating from noise, PGGAN \cite{karras2017progressive} achieves $1024^2$ resolution for the first time, and then subsequent work \cite{brock2018large} improves the sample variety and quality. Later on, a style-based generator \cite{karras2018style} is represented to achieve conditional high-resolution generation. Wang \etal \cite{wang2018high} make progress on high-resolution interactive image editing via semantic label maps. Yin \etal \cite{yin2018instance} focus on high-resolution facial attribute editing and propose a scale-invariant approach based on geometric flow. Compared with existing work, the most significant difference is that we aim at achieving the translation of any two visual domains in a multi-domain translation task by training one single generator.

The notion of \textbf{Knowledge distillation} (KD) is originally proposed by Hinton \etal \cite{hinton2015distilling}. The main purpose is to transfer the learned knowledge from a large powerful teacher network to a small compact student one. To narrow the performance gap between student and teacher networks, embedding matching \cite{romero2014fitnets}, attention map matching \cite{zagoruyko2016paying}, Gramian matrix matching \cite{yim2017gift} methods are successively proposed. A new perspective of KD is brought by the finding in Born-Again Network (BANs) \cite{furlanello2018born}: Training a student net with identical model structure can bring a significant outperformance than the teacher. Inspired by BANs, we introduce KD methods to train a student net that has enhanced model capacity and deals with a higher resolution than the teacher. In this work, we extend KD to the area of adversarial image generation and propose the inherited adversarial loss.

Representation learning methods aim at seeking informative representation with desired properties, including compactness \cite{gretton2012kernel}, disentanglement \cite{gonzalez2018image}, independence \cite{brakel2017learning}, \etc. \textbf{Mutual information} (MI) is a reliable measure of statistical dependence that depicts the informativeness and diversity between random variables, but how to compute the MI of high-dimensional data is an intractable problem. Recently, a very promising general-purpose solution \cite{brakel2017learning} that resorts to training a deep neural MI estimator is proposed. Subsequent works \cite{belghazi2018mine, hjelm2018learning} make further improvements and verify the effectiveness on an extensive range of applications. In this work, we explore a supervised training approach to replace the joint adversarial training for the estimator.

\section{Approach}

Given $N$ different visual domains $\{X_{i}\}_{i=1}^{N}$, our goal is to learn the mappings among these domains from training samples. We assume that:
\begin{equation}
\begin{split}
\label{e1}
\forall~i,j \in \{1, 2, \cdots, N\},i\neq j,~\exists~f_{ij}:X_{i}\mapsto X_{j},
\end{split}
\end{equation}
where $f_{ij}$ is a bijection. Furthermore, we assume that the training data is unpaired. We make these assumptions based on the conditions of a wide range of translation tasks, including facial attribute transfer \cite{choi2018stargan}, facial animation \cite{pumarola2018ganimation}, \etc. In this paper, we consider that all the visual domains consist of human faces. We mainly study the situation where the visual domains consist of extremely high-dimensional data, \ie, images at a resolution of $1024^2$ in our experiments.

\subsection{Biphasic adversarial learning}

We first describe the core idea of our biphasic learning, which is illustrated in Fig. \ref{f2}. We aim at employing a single conditional generator to learn the multi-domain mapping function. In the initial phase, we train a network $G_{1}$ to take the input image $\bm{x}$ at $256^2$ resolution and the target domain label $c$ to achieve the low-resolution translation. In the meantime, the corresponding discriminator, $T$, provides adversarial supervision for $G_{1}$ by learning to discriminate the real and the synthesized images. Without loss of generality, we adopt the vanilla adversarial loss function of GAN:
\begin{equation}
\label{e2}
\mathcal{L}^{G}_{adv}=\mathbb{E}_{\bm{x},c}[-\log(T(G_{1}(\bm{x},c))],
\end{equation}
where the superscript on the loss item indicates that the loss belongs to the generator or the discriminator. Correspondingly, the adversarial loss for $T$ is formulated as:
\begin{equation}
\begin{split}
\label{e3}
\mathcal{L}^{D}_{adv} = ~&\mathbb{E}_{\bm{x}}[-\log(T(\bm{x}))]~~+ \\
                        &\mathbb{E}_{\bm{x},c}[-\log(1-T(G_{1}(\bm{x},c))]. 
\end{split}
\end{equation}

In the initial phase, we employ the standard adversarial learning scheme and build our backbone generation model. We emphasize our major contribution is exploring an effective method to achieve the inheritance and enhancement of model capacity in the staged training process, which is reflected in the enhancing phase. In this phase, our model is trained for full-resolution image generation, and we replace the loss function with the proposed inherited adversarial loss. To adopt the increase of the spatial resolution, we attach additional neural layers $G_{2}$ to $G_{1}$, proposing the enhanced generator $G$. Whereas we train a new discriminator $D$ with similar architecture but more layers under the supervision of the former discriminator $T$, which is regarded as the teacher net.

Recall that recent work \cite{furlanello2018born} in Knowledge Distillation (KD) \cite{hinton2015distilling} pointed out that benefiting from the dark knowledge, the student net can even achieve a better performance than the teacher net. However, \cite{furlanello2018born} assume that the training data distribution is static, \ie, the teacher and the student learn from the same data distribution that remains unchanged across the training process. This assumption does not hold in our adversarial learning situation because $D$ draws fake data from $G(\bm{x},c)$, which varies with the change of its parameters. Actually, replacing the data label with the prediction of $T$ in the second term of Eq. \ref{e3} violates the principle of adversarial training: if a sample is misclassified by $T$, the label replacement will advocate $D$ neglect this mistake. Therefore, in Eq. \ref{e3}, the guidance introduced from KD is only feasible for the first term because $D$ still draws real data from a static data distribution. In summary, the inherited adversarial loss for our discriminator is formulated as:

\begin{small}
\begin{equation}
\begin{split}
\label{e4}
\mathcal{L}^{D}_{adv} &= \mathbb{E}_{\bm{x},c}[-\log(1-D(G(\bm{x},c))]~+ \\ 
                      &\! \mathbb{E}_{\bm{x}}[-T(\bm{x})\log(D(\bm{x})) - (1-T(\bm{x}))\log(1-D(\bm{x}))].
\end{split}
\end{equation}
\end{small}

In Equation \ref{e4}, the supervision for classifying the real data is provided by the teacher net instead of the ground truth label. Learning the dark knowledge distilled from the initial phase, our network is less sensitive to the noise in the training data.

However, we have not got a complete distillation method for GANs yet. The power of $T$ is not been fully utilized because Eq. \ref{e4} discards the knowledge about the fake data. To address this issue, we design a sample importance weighting method for our $G$. Formally, the inherited adversarial loss for the generator is:

\begin{equation}
\mathcal{L}_{adv}^{G}=\mathbb{E}_{\bm{x},c}[-\frac{\omega(T(\bm{x}))}{2}\cdot\log(D(G(\bm{x},c))],
\end{equation}
where $\omega(T(\bm{x}))=1+\frac{T'(\bm{x})-min(T'(\bm{x}))}{max(T'(\bm{x}))-min(T'(\bm{x}))}$, and we define $T'(\bm{x})=1-T(\bm{x})$. The first constant item in our weighting factor stands for the weight inherited from the ordinary adversarial training; The second item ranges between 0 and 1 and attaches more importance to those samples that are not good at deceiving the teacher net. Equipped with our importance weighting scheme, the generator combines the guidance from the data label and the prediction of the teacher net.

\subsection{Mutual perceptual information maximization}

Although applying the proposed approaches above effectively guides our network to transfer large-scale images with high quality, there is still one issue remained unsolved: the consistency of these irrelevant attributes. For instance, if we want to change the hair color of a given source face image, on the one hand, the color should be changed and the visual realism should be preserved; On the other hand, we expect that the difference in the attributes except for the hair color is imperceptible. However, calculating the adversarial loss alone does not consider the correspondence between the input and the output images. Therefore, we advocate maximizing mutual information (MI) to maintain the consistency between the input and the output.

The mutual information of high-dimensional data is notoriously difficult to compute, but recently brand new approaches \cite{brakel2017learning,belghazi2018mine,hjelm2018learning} based on GANs have been explored. They estimate MI by training a deep neural estimator jointly with the generator and the discriminator. Different from their methods, we employ a classifier pre-trained on external datasets in the supervised manner. Since we focus on generating human faces in our experiments, we resort to a face recognizer $T_{\omega}$ trained on publicly available face datasets as the estimator. $T_{\omega}$ measures image similarity in the perceptual level, so our proposed MI maximization is referred to as mutual perceptual information maximization. Concretely, given a pair of images, $\bm{x}_1$ and $\bm{x}_2$, the similarity is calculated as:

\begin{equation}
\label{e6}
T_{\omega}(\bm{x}_1,\bm{x}_2)=\text{\ensuremath{\Vert}}\phi(\bm{x}_1)-\phi(\bm{x}_2)\Vert_{2},
\end{equation}
where $\phi(\cdot)$ denotes the extracted identity representation obtained by the second last fully connected layer within $T_{\omega}$ and $\Vert\cdot\Vert_{2}$ means the vector 2-norm. Then our mutual perceptual information loss is defined as:

\begin{equation}
\label{e7}
L^{G}_{mp}=-\widehat{\mathcal{I}}(\bm{x},G(\bm{x},c);T_{\omega}),
\end{equation}
where $\widehat{\mathcal{I}}$ represents the mutual information estimation funciton. Specifically, we adopt Noise-Contrastive Mutual Infomation Estimation (infoNCE) \cite{oord2018representation}. Combining Eq. \ref{e6}, Eq. \ref{e7}, and infoNCE, our mutual perceptual information is measured as: 

\begin{small}
\begin{equation}
\label{e8}
\begin{split}
\widehat{\mathcal{I}}(\bm{x},G(\bm{x},c);T_{\omega})&=\mathbb{E}_{\bm{x},c}\Big[T_{\omega}(\bm{x},G(\bm{x},c))~- \\
\mathbb{E}_{\tilde{\bm{x}},c}&\left[\log\sum\nolimits_{\tilde{\bm{x}}}e^{T_{\omega}(\bm{x},G(\tilde{\bm{x}},c)}\right]\Big],
\end{split}
\end{equation}
\end{small}
where $\tilde{\bm{x}}$ denotes a data batch resampled from the training set. During the two training phases, we both calculate $L^{G}_{mp}$ in the same manner, \ie, the form of $L^{G}_{mp}$ does not change across our biphasic learning process.

\subsection{Biphasic regularization}

We also integrate several regularization items into our total loss function. These items have been proved to be beneficial for normal-resolution image generation tasks. In this work, we adapt them to the biphasic form to cope with high-resolution conditions.

We train an auxiliary network, $D^{uv}$, to predict the UV correspondence field (the UV field for short) of the input face. The UV field is a powerful representation that depicts the geometric information of the human face. Compared with other geometric guidances for face generation, \eg, landmarks and facial parsing maps, the UV field has two significant advantages: 1) The input face is related to an implicit facial texture map that indicates its identity, and this relationship is pixel-wise; 2) 3D geometric prior is subtly integrated by the UV field. Following \cite{ cao2018learning}, we construct the ground truth UV field via fitting a 3D Morphable Model \cite{paysan20093d} through the Multi-Features Framework \cite{romdhani2005estimating}. In the initial stage, $D^{uv}$ is trained to correctly predict the UV field of the real faces, whereas $G$ is trained to make $D^{uv}$ think the transformed faces have the same UV field with its corresponding real faces. Concretely, we optimize:

\begin{align}
L_{geom}^{D^{uv}}&=\mathbb{E}_{\bm{x}}\Vert D^{uv}(\bm{x})-\bm{uv}\Vert_{1}, \\
L_{geom}^{G}&=\mathbb{E}_{\bm{x},c}\Vert D^{uv}(G_{1}(\bm{x},c))-\bm{uv}\Vert_{1},
\end{align}
where $\Vert \cdot \Vert_{1}$ denotes calculating the mean of the element-wise absolute value summation of a matrix. In the enhancing phase, we fix the parameters of $D^{uv}$, \ie, $L_{geom}^{G}$ is optimized with the guidance rather than the competition of $D^{uv}$. The reason is that normal-resolution images are already sufficient for estimating facial geometric information. Therefore, keeping training $D^{uv}$ in the enhancing phase is of no benefit. We find that applying our biphasic UV loss is a simple yet effective manner to improve the experimental performance. Note that in our following experiments, if we consider to alter those attributes that change the original facial geometric information, \eg, changing facial expressions,  the UV loss is excluded.

Furthermore, we introduce an auxiliary classifier, $D^{a}$, to predict the facial attribute. Similar with the UV loss, $D^{a}$ and $G$ are trained jointly in an adversarial manner in the initial phase, and $D^{a}$ is transformed into a fixed network to supervise $G$ in the enhancing phase. $L_{attr}^{G}$ and $L_{attr}^{D^{a}}$, \ie, the attribute loss for $G$ and $D^{a}$, are measured by the cross-entropy loss function. The attribute loss guides our network on the attribute translation in semantic level, ensuring that the target attribute is imposed on the output.

\begin{figure*}[h!]
\captionsetup{aboveskip=0pt}
\begin{center}
\includegraphics[width=\textwidth]
{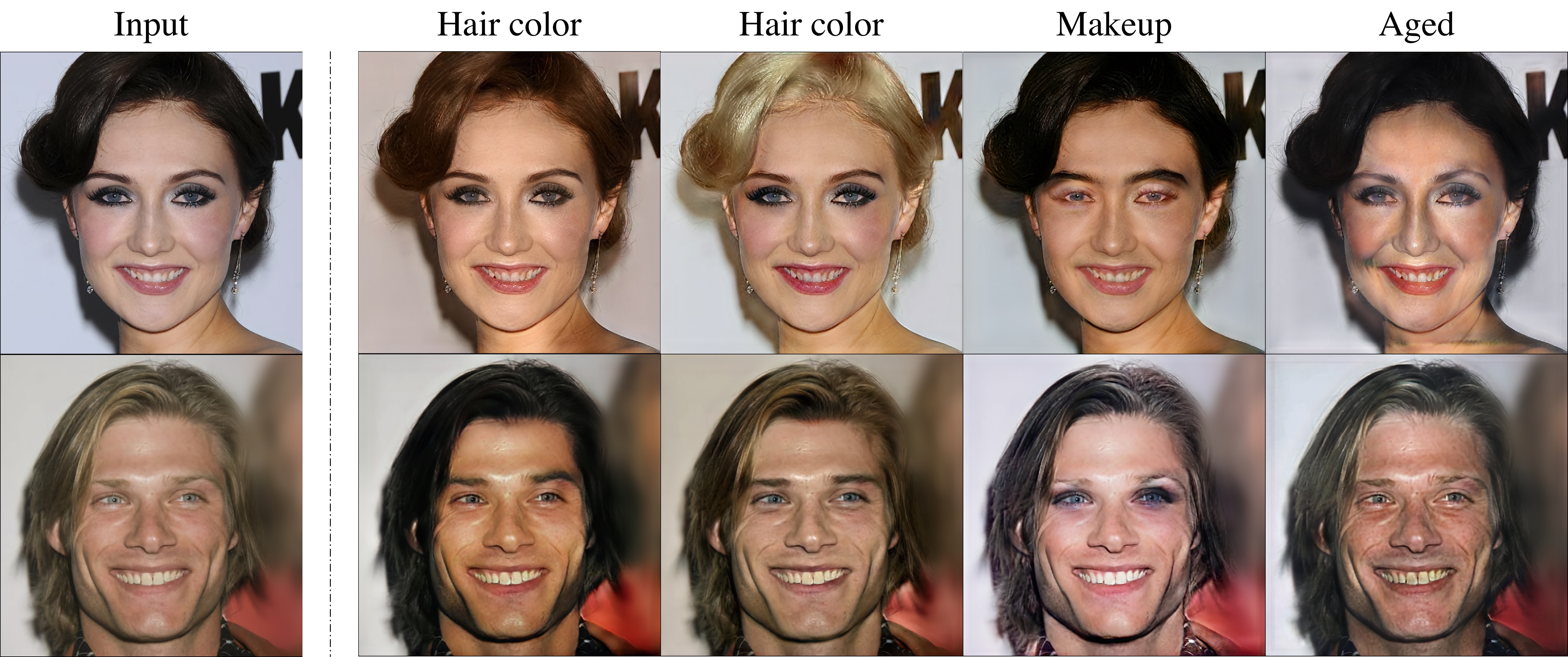}
\end{center}
\caption{Facial attribute transfer results on CelebA-HQ. The first column shows the input image, and the rest four columns show transformed results. Due to the limitation of submission size, we have to downsample this figure. Please zoom in for better visualization.}
\label{f3}
\end{figure*}

\subsection{Implementation details}

The structure of our network is mainly adapted from existing work. We inherit the model structure in StarGAN \cite{choi2018stargan} to build our $G_{1}$ and $T$ in the initial phase. In the enhancing phase, the structure for $D$ remains the same except for some minor modifications to adapt the higher input resolution. Whereas for the generator, a fully convolutional backend network used in \cite{ledig2017photo} is added to produce the enhanced results. Detailed descriptions are given in Appendix B. We denote our network as Biphasic Learning GAN (BiL-GAN) in the following parts. 

In our experiment, we replace the vanilla GAN objective function with LSGAN \cite{mao2017least}. In summary, our total loss functions are formulated as: 

\begin{align}
L^{G}&=\ddot{L_{adv}^{G}}+L_{mp}^{G}+L_{geom}^{G}+L_{attr}^{G}, \\
L^{D}&=\ddot{L_{adv}^{D}}+(L_{geom}^{D^{uv}})+(L_{attr}^{D^{a}}),
\end{align}
where the items in brackets do not appear in the enhancing training phase and the items with two dots on top are calculated differently across the two phases. We set all the weight factors of the loss items to $1$.

We optimize the parameters of our model by Adam \cite{kingma2014adam} optimizer with $\beta_1 = 0.5$, $\beta_1 = 0.999$, and a learning rate of 1e-4. For data augmentation, we flip the input images horizontally with a probability of 0.5. The batch size is set to $64$ for all experiments. We train our BiL-GAN on 8 Tesla V100 GPUs for approximately one week. The Appendix is available here\footnote{\url{https://drive.google.com/file/d/17t3YwB1OMYoIMaB-8XU55gxzRfM-p7qP/view?usp=sharing}}. The source code will be available soon.

\section{Experiments}

\subsection{High-resolution image-to-image translation}
To achieve image-to-image translation at $1024^2$ resolution, we train our BiL-GAN on CelebA-HQ \cite{karras2017progressive}, which is the most widely used face dataset for high-quality image synthesis at present. We randomly choose 300 images from the total 30,000 ones for testing and use the rest part as the training set. We consider image-to-image translation in seven different domains: hair color (black, blond, brown), makeup (with/without), and age (young/old).

Our BiL-GAN is able to alter the involved attributes to any specified type, and we do not apply any post process, \eg, warping the transformed facial region to the original one to make the background identical \cite{li2018beautygan}. Some visual samples are shown in Fig. \ref{f1} for preview and the attribute translation results are shown in Fig. \ref{f3}. Please refer to Appendix A for more samples and transfer types. We argue that the visual quality of the transformed images should be judged by the following three points: (1) the overall realism, (2) the transfer quality of attribute(s), (3) the preservation of the original identity and the consistency of the uninvolved attribute(s). We can observe that our method can precisely change the target attribute and faithfully preserves the character of the original input, including expression, background, \etc. Furthermore, we find most transformation results on hair color are very difficult for a human to distinguish even with unlimited observation time.

\begin{table}
\captionsetup{aboveskip=0pt}
\begin{center}
\scalebox{0.9}
{
	\begin{tabular}{lcccc}
		\toprule
		\multicolumn{1}{c}{Method} & Aging & Makeup & Hair color \\
		\midrule
		Ours $>$ IcGAN \cite{perarnau2016invertible} & 90.7\% & 89.6\% & 94.3\% \\
		Ours $>$ CycleGAN \cite{zhu2017unpaired} & 86.3\% & 85.5\% & 89.2\% \\
		Ours $>$ StarGAN \cite{choi2018stargan} & 73.7\% & 71.9\% & 72.4\% \\
		\bottomrule
	\end{tabular}		
}
\end{center}
\caption{User study results across different attributes on CelebA-HQ. The percentages of the cases where the user prefers our results are listed above.}
\label{t1}
\end{table}

We systematically compare our results with existing methods in all aspects. Concretely, we compare with IcGAN \cite{perarnau2016invertible}, CycleGAN \cite{zhu2017unpaired}, and StarGAN \cite{choi2018stargan}. It is notable that these methods are all designed for facial attribute translation on CelebA \cite{liu2015deep} which only consists of normal-resolution images. We have tried to train these methods with $1024^2$ resolution images directly, but they produced unrealistic results with severe degradation (cf. Appendix C). Thus we downsample the images in CelebA-HQ to a resolution of $256^2$ and use them to retrain the models of these approaches. However, the performance drop of existing methods is inevitable because of the reduction of training data (CelebA-HQ is only a subset of CelebA). In the following comparisons, we also downsample our results from $1024^2$ to $256^2$ and calculate quantitative evaluation results.

\begin{table}
\captionsetup{aboveskip=0pt}
\begin{center}
\scalebox{0.9}
{
	\begin{tabular}{cccc}
    \toprule
    Method & FID   & MS-SSIM & CLAS Error \\
    \midrule
    IcGAN \cite{perarnau2016invertible} &  24.16     &  0.8072     &22.17\%  \\
    CycleGAN \cite{zhu2017unpaired} &  25.86     &  0.8384     & 14.86\% \\
    StarGAN \cite{choi2018stargan} &  19.75     &  0.8509     & 9.37\% \\
    BiL-GAN (ours) &  \textbf{14.93}     &  \textbf{0.8563}     & \textbf{5.49\%} \\
    \midrule
    \small{single phase learning}  &  21.38     &  0.8372     & 11.70\%  \\
    \small{progressively learning}  &  17.94     &  0.8511     & 8.39\%  \\
    \small{ordinary regularization} &  16.36     &  0.8605     & 6.93\%  \\
    \small{$w/o$ mutual Info loss} &  19.15     &  0.8541     & 10.48\%  \\
    \small{$w/$ MINE loss} &  17.92     &  0.8539     & 7.34\%  \\
    \small{$w/$ cycle loss} &  18.56     &  0.8420     & 9.28\%  \\
    \bottomrule
    \end{tabular}
}
\end{center}
\caption{Comparisons of Fr\'echet inception distance (FID), multi-scale structural similarity (MS-SSIM), and the attribute classification (CLAS) error on the CelebA-HQ dataset. The upper part shows comparisons with existing methods, and the lower part shows comparisons with our proposed variations for ablation study.}
\label{t2}
\end{table}

We conduct a user study to evaluate the visual quality of samples. For each face in our testing set, we produce 8 different transformed images, including 5 ones with single altered attribute (hair color, makeup, or age), 3 ones with two altered attributes. Hence, we get 2,400 transformed images. Users are asked to compare two transformed images that are controlled by the target domain label but produced by two different methods, and then judge which one is better. We record the percentage of the cases where the user prefers our results. Thus, a higher percentage indicates better performance of our method and 50\% means that the user thinks two methods are equal. Different attributes are counted separately and the percentages are averaged over all participated users. Synthesized images with two altered attributes are counted by both attributes. The experimental results of the user study are summarized in Table \ref{t1}, which demonstrates the superiority of our BiL-GAN.

We also calculate the Fr\'echet inception distance (FID) \cite{heusel2017gans}, multi-scale structural similarity (MS-SSIM) \cite{odena2017conditional}, and the attribute classification error to make objective comparisons. FID has been recently proposed to reveal the performance of general image generation tasks. In our experiment, we follow \cite{miyato2018cgans} and compute FID between the real and the synthesized faces. Since lower FID score indicates that the Wasserstein distance between two distributions is smaller, the method with the lowest FID is considered to be the one that produce the most realistic samples. MS-SSIM values range between 0 and 1, and a higher value indicates stronger perceptual similarity. We compute the MS-SSIM between the transformed images and the inputs to evaluate the ability to keep overall consistency. In the meantime, we employ a pre-trained facial attribute classifier \cite{liu2015deep} to reflect the effectiveness of transformation. The synthesized samples are supposed to be classified to the classes that accord with their corresponding target domain labels. In summary, we expect that a good method has low FID, MS-SSIM, and classification error rate. We report these objective performance measurements in the upper part of Table \ref{t2}. It can be observed that our BiL-GAN outperforms other approaches by a large margin. Our FID and classification error are markedly lower than the second-best method, which clearly indicates our improvements in visual realism and attribute transfer.

\begin{figure}[!t]
\captionsetup{aboveskip=0pt}
\begin{center}
\includegraphics[width=0.5\textwidth]
{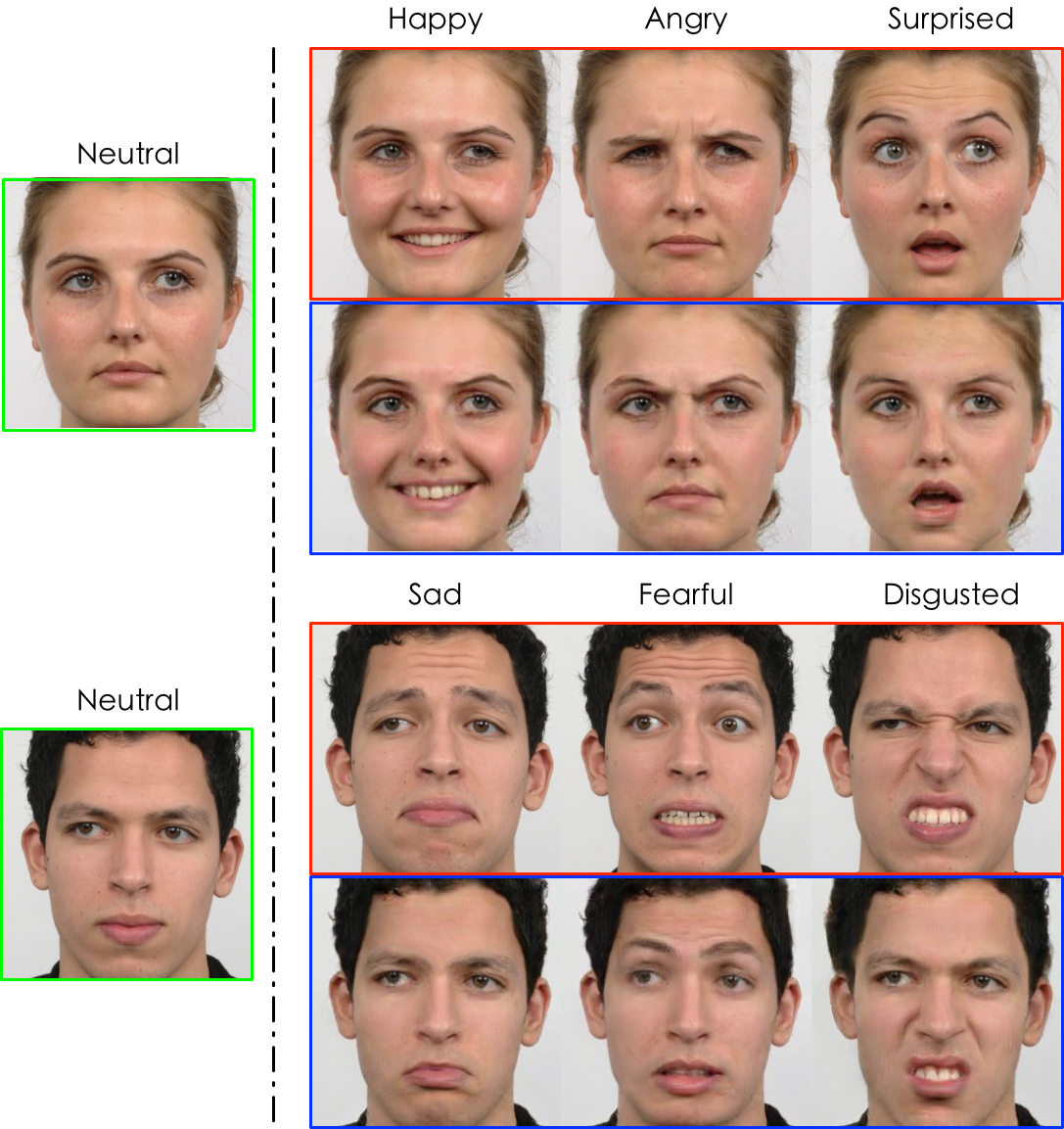}
\end{center}
\caption{Facial animation results on RaFD. Given the input (in the green box), our BiL-GAN is able to produce diverse results (in the blue box) with specified facial expression. The ground truth is also shown as a reference (in the red box).}
\label{f4}
\end{figure}

\subsection{Ablation study}

In this section, we investigate the contributions of our biphasic learning and mutual perceptual information regularization by ablation study. To this end, we design a series of variations and report their performances in the lower part of Table \ref{t2}. Our analyses are listed below:

(1)	We train a network totally without any biphasic schemes (denoted as ``single phase learning'' in the table). The network is simply trained by following the idea of classical adversarial training \cite{goodfellow2014generative}. Through comparisons, we observe that single phase learning leads to much inferior performance. Besides, there are significant flaws in the produced samples. Our observation accords with the conclusions in \cite{karras2017progressive, wang2018high}: ordinary single stage adversarial training shows a limited capacity to generate high-resolution images.

(2)	We train a network without both the inherited adversarial loss and biphasic regularization items (denoted as ``progressively learning''), and a network just without biphasic regularization items (denoted as ``ordinary regularization''). We also add the resolution transition process \cite{karras2017progressive} for the former variation since it reflects the idea of PGGAN. Through comparisons, we can see the superiority of the proposed biphasic learning over the mainstream progressively growing method. Thanks to our supervision transition scheme and the inherited adversarial loss, BiL-GAN has a considerably lower FID score. Besides, compared with ``ordinary regularization'', BiL-GAN further lowers the FID score and the classification error at the cost of a slightly higher MS-SSIM.

(3) We train a network without our mutual perceptual loss (denoted as ``$w/o$ mutual Info loss''), a network that maximizes mutual information as described in MINE \cite{belghazi2018mine} (denoted as ``$w/$ MINE loss''), and a network that replaces mutual information loss with the cycle consistency loss \cite{zhu2017unpaired} (denoted as ``$w/$ cycle loss''). We design this group of variations to verify the effectiveness of the proposed mutual perceptual information loss. It can be observed that the performances of ``$w/o$ mutual Info loss'' and ``$w/$ cycle loss'' are inferior, and BiL-GAN also outperforms the variation that employs adversarial mutual information loss, demonstrating that our pre-trained neural estimator provides more effective supervision. 

\begin{figure}[!t]
\captionsetup{aboveskip=0pt}
\begin{center}
\includegraphics[width=0.5\textwidth]
{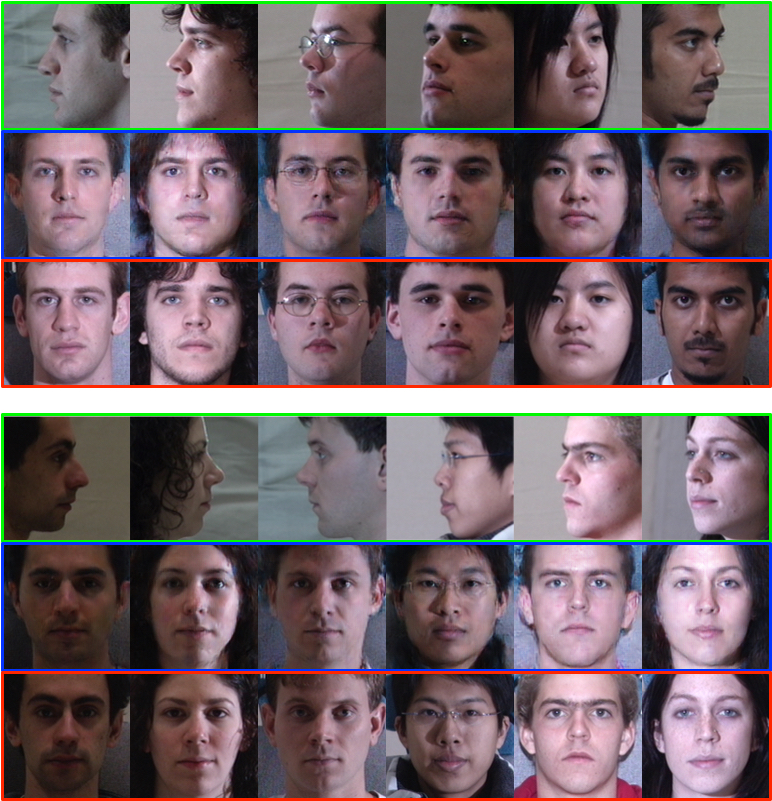}
\end{center}
\caption{Face frontalization results on Multi-PIE. Given the input across different poses (in the green box), our BiL-GAN produces frontalized results (in the blue box). The ground truth is also shown as a reference (in the red box).}
\label{f5}
\end{figure}

\subsection{Image-to-image translation for specific tasks}

In the discussions above, we mainly focus on addressing general semantical image-to-image translation problem. Whereas the image-to-image translation can be divided into many specific tasks. To demonstrate our methods can be generalized to synthesis tasks in various conditions, we test our biphasic learning framework on two specific facial attribute transfer tasks, \ie, facial animation and face frontalization. Accordingly, we make three minor modifications on our BiL-GAN: we remove some up-sampling layers to adjust the spatial size of the output because following experiments are conducted on currently available normal-resolution datasets; we introduce pixel loss calculated with the ground truth because paired data is available; we remove the UV loss because the assumption that the facial geometric information should remain unchanged does not hold.

We conduct the facial animation experiment on RaFD \cite{langner2010presentation}, which consists of 67 identities with 8 binary labels for facial expressions in three gaze directions. We follow the \cite{choi2018stargan} to preprocess and split the dataset and produce $256^2$ results. Synthesized samples are shown in Fig. \ref{f4}. We observe that our results have appealing visual realism and faithfully preserve the identity and gaze direction. Besides, another interesting finding is that our results have the same facial expression with the ground truth, but they are different in the pixel domain.

\begin{table}[!t]
\captionsetup{aboveskip=0pt}
\begin{center}
\scalebox{0.8}
{
	\begin{tabular}{ccccccc}
		\toprule
		Method& \(\pm 15^{\circ}\) & \(\pm 30^{\circ}\) & \(\pm 45^{\circ}\) & \(\pm 60^{\circ}\) & \(\pm 75^{\circ}\) & \(\pm 90^{\circ}\)\\
		\midrule
		DR-GAN \cite{tran2017disentangled}& 94.9 & 91.1 & 87.2 & 84.6 & - & - \\
		FF-GAN \cite{yin2017towards}& 94.6 & 92.5 & 89.7 & 85.2 & 77.2 & 61.2 \\
		TP-GAN \cite{huang2017beyond}& 98.7 & 98.1 & 95.4 & 87.7 & 77.4 & 64.6 \\
		CAPG-GAN \cite{hu2018pose} & 99.8 & 99.6 & 97.3 & 90.3 & 83.1 & 66.1 \\
		PIM \cite{zhaotowards} & 99.3 & 99.0 & 98.5 & 98.1 & 95.0 & 86.5 \\
		3D-PIM \cite{zhao20183d} & 99.6 & 99.5 & 98.8 & 98.4 & 95.2 & 86.7 \\
		HF-PIM \cite{cao2018learning} & 99.9 & 99.9 & 99.9 & 99.1 & 96.4 & 92.3 \\
		\midrule
		Light CNN \cite{he2017learning}& 98.6 & 97.4 & 92.1 &62.1 & 24.2 & 5.5 \\
		\textbf{BiL-GAN(Ours)} & \textbf{99.96} & \textbf{99.94} & \textbf{99.93} & \textbf{99.90} & \textbf{96.92}& \textbf{93.11} \\
		\bottomrule
	\end{tabular}
}
\end{center}
\caption{Comparisons on rank-1 recognition rates (\%) across views under Multi-PIE Setting 2.}
\label{t3}
\end{table}

Our face frontalization experiment is conducted on Multi-PIE \cite{gross2010multi}, which consists 13 poses within $\pm {90^{\circ}}$ of 337 subjects. Those profiles with extreme poses (${75^{\circ}}$ and ${90^{\circ}}$) are very challenging for cross-view face synthesis. We train BiL-GAN to frontalize the profiles across all poses at $128^2$ resolution and use the first 200 subjects for training and the rest 137 for testing. Frontalization results are visualized in Fig. \ref{f5}. Face frontalization methods are mainly evaluated by the performance of face recognition. To this end, existing works \cite{huang2017beyond,yin2017towards,hu2018pose} have explored the ``generation via recognition'' framework to calculate the recognition rate, making a standard manner for comparisons among different methods. Therefore, we follow \cite{cao2018learning} and employ a pre-trained Light CNN \cite{he2017learning} to calculate our face recognition rate and make comparisons with existing methods, as shown in Table \ref{t3}. Please refer to Appendix B for more details about the recognition rate calculation. We can see that our approach further improves the recognition rate in extreme poses, indicating that our model not only produces results with high visual quality but also preserves the identity information.

\section{Conclusion}

This paper proposes a novel training method for GANs, namely biphasic learning, to improve the performance of image-to-image translation. We apply our approach to a series of applications, including high-resolution and normal-resolution, supervised and unsupervised, general and specified translation tasks. As a result, our model produces visually appealing samples that outperform competing approaches in both objective and subjective evaluations. However, we discover that the quantity and quality of the training data still limit the performances to a great extent (cf. Appendix D). Fortunately, recently proposed FFHQ dataset \cite{karras2018style} offers human face images with higher quality and wider variation. In the future, we will explore further improvements with larger datasets.

\clearpage

{\small
\bibliographystyle{ieee}
\bibliography{egpaper_for_review}
}

\end{document}